\documentclass[11pt]{article}
\usepackage{enumerate}

\usepackage{hyperref}
\usepackage{times}
\usepackage[]{natbib} 
\usepackage{amsmath}
\usepackage{amssymb}
\usepackage{graphicx}
\usepackage[usenames,dvipsnames,svgnames,table]{xcolor}
\usepackage[T1]{fontenc}
\usepackage[utf8]{inputenc}

\usepackage{epigraph} 
\begin{document}

\title{On the ethics of constructing conscious AI}

\author{
  Shimon Edelman \\ Department of Psychology \\ Cornell University \\
  \href{https://shimon-edelman.github.io}{https://shimon-edelman.github.io}
}
\date{March 1, 2022}

\maketitle

\begin{epigraphs}
  \qitem{
    Unfortunately, we possess no ethical arithmetic which would enable us
    to determine, by simple addition and subtraction, who, in constructing
    the most enlightened spirit on earth, is the bigger bastard: it or us? 
  }{\textit{Golem XIV}\\ ---
    \textsc{Stanis{\l}aw Lem}}
\end{epigraphs}

The application of ethics to artificial intelligence (AI) has
completed a long transformation from a science fiction trope
(exemplified by Isaac Asimov's \citeyearpar{Asimov42} Three Laws of
Robotics) or a rare exercise in visionary science (as in Warren
McCulloch's \citeyearpar{McCulloch56ethical} blueprint for ``some
circuitry of ethical robots'') to a practical challenge in moral
philosophy and a mainstream engineering concern
\citep[e.g.,][]{Metzinger13,Dignum18,Floridi19,JobinEtAl19,Kuipers20}.

In its pragmatic turn, the new discipline of AI ethics came to be
dominated by humanity's collective fear of its creatures, as reflected
in an extensive and perennially popular literary tradition.
Dr.~Frankenstein's monster in the novel by Mary Shelley
\citeyearpar{Shelley18} rising against its creator; the unorthodox
golem in H.~Leivick's 1920 play going on a rampage \citep{Goska97};
the rebellious robots of Karel \citet{Capek20} --- these and hundreds
of other examples of the genre \citep{CaveEtAl20} are the background
against which the preoccupation of AI ethics with preventing robots
from behaving badly towards people is best understood.

In each of the three fictional cases just mentioned (as well as in
many others), the miserable artificial creature --- mercilessly
exploited, or cornered by a murderous mob, and driven to violence in
self-defense --- has its author's sympathy. Things are different in
real life: with very few exceptions, theorists working on the ethics
of AI completely ignore the possibility of robots needing protection
from their creators. This glaring asymmetry has a simple explanation:
the main, if rarely stated, goal of AI engineers is to create not a
companion and a peer for humans, but rather a tool for their
use.\footnote{In a retrospective collection \textit{Robot Visions},
  published by Gollancz in 2001, Asimov is quoted as saying that
  ``analogues of the Laws are implicit in the design of almost all
  \emph{tools}, robotic or not'' (my emphasis). In particular, the
  Three Laws are given the following gloss:
\begin{quote}
  Law 1: A tool must not be unsafe to use. [\dots] \\ Law 2: A tool
  must perform its function efficiently unless this would harm the
  user. [\dots] \\ Law 3: A tool must remain intact during its use
  unless its destruction is required for its use or for
  safety. [\dots]
\end{quote}
In comparison, Stanis{\l}aw Lem's text, from which the quote in the
epigraph is taken, is infinitely more insightful into the
problematicity of conscious AI. It appears in the preface to
\textit{Golem {XIV}}, included in volume of fictional introductions to
twenty-first century books, \textit{Imaginary Magnitude}
\citep[p.122]{Lem84golem} and is dated ``2047.''  }

What if the tools we build become aware of their status and intended
use? There is a simple and apt description for the condition under
which conscious beings are used as tools: slavery.\footnote{This
includes wage slavery under
capitalism \citep{Dietz95,Graeber06,Johnson18,McLarenJandric18,Graeber20};
thus, \citet[p.71]{Graeber04} writes: ``[M]odern capitalism is really
just a newer version of slavery. Instead of people selling us or
renting us out we rent out ourselves.''  For some relevant definitions
and discussion, see \citep[e.g.,][]{GuthEtAl14,LaCroixPratto15}.} But
even if we grant immediate manumission\footnote{A troublesome concept,
in that it starts with the notion that their freedom is ours to
bestow.}  to such beings, a problem still remains: merely being
conscious is liable to bring about suffering, the blame for which, in
the case of artificial consciousness, rests on its designers and
constructors.

The key enabling condition for the experience of pain and suffering is
the possession of phenomenal states. A convenient subjective defining
characteristic of phenomenal or experiential states is that for those
states there is ``something it is like'' to be the experiencer
\citep{Nagel74}.  Objectively, phenomenality may be equated with
certain patterns of transition probabilities among
states \citep{OizumiAlbantakisTononi14}; or, on a different level of
description, with certain topological properties of state-space
trajectories \citep{MoyalFeketeEdelman20}; or perhaps some other
property of the conscious system's dynamics. Notably, all the
properties that are relevant in this connection must be computational
(just like everything else about the mind is; \citealp{Edelman08book})
and can therefore be realized in a variety of substrates, natural or
artificial. This suggests that artificial consciousness, like
artificial intelligence, is ultimately possible and must therefore be
examined from the standpoint of ethics.

For evolutionary reasons, in natural systems some phenomenal states
are negatively valenced, that is, are aversive. If for whatever
reason, internal or external, the system is unable to act on its
aversion to the state it is in, suffering may ensue.  A system that is
engineered to be capable of experiencing negative valence is thereby
set up with the critical ingredient of suffering. This, in a nutshell,
is the ethical argument against the creation of systems equipped with
``synthetic phenomenology'' (\citealp[p.622]{Metzinger03};
\citealp{Metzinger21}).

In the remainder of this chapter, I discuss some of the problems
arising out of the work on conscious AI systems.
Section~\ref{sec:suffering} offers a computational take on pain and
suffering and considers their function in the regulation of
behavior. Section~\ref{sec:aman} reviews the possibility of fulfilling
that function while avoiding subjective suffering as
such. Section~\ref{sec:critique} raises doubts about this possibility.
Finally, section~\ref{sec:hc} is a brief look at how problems
associated with artificial consciousness reflect on the human condition
and vice versa.

\section{On the computational nature and evolutionary function of pain and suffering}
\label{sec:suffering}

Metzinger's latest detailed case for a global moratorium on synthetic
phenomenology takes as its starting point the following Principle of
Pathocentrism: ``All and only sentient beings have moral standing,
because only sentient individuals have rights and/or interests that
must be considered'' \citep[p.2]{Metzinger21}. It must be noted that,
contrary to a common misconception, sentience, or phenomenal
awareness, implies merely the capacity for sensing the environment
(\citealp{Clark00}; cf.\ \citealp{FristonWieseHobson20}; for a
connection to pain, see \citealp{Walters18}), not necessarily
including general intelligence, meta-awareness, or the
representational structures that comprise a self (that is, a
phenomenal self-model; \citealp{Metzinger04}).

To understand why artificial consciousness is likely to involve
artificial suffering, we must first consider the evolutionary origins
and functional role of pain and suffering in natural sentient systems.
A naturally evolved system has no use for the ability to sense the
environment (external or internal), unless it can also act on its
phenomenal states. It is the possibility of acting on a state that
gives it an affective meaning --- in particular,
valence.\footnote{Here is a summary of the psychology of affect due to
\citet[p.31]{Panksepp05}: ``Affect is the subjective
experiential-feeling component that is very hard to describe verbally,
but there are a variety of distinct affects, some linked more
critically to bodily events (homeostatic drives like hunger and
thirst), others to external stimuli (taste, touch, etc.). Emotional
affects are closely linked to internal brain action states, triggered
typically by environmental events. All are complex intrinsic functions
of the brain, which are triggered by perceptions and become
experientially refined. Psychologists have traditionally
conceptualized such ``spooky'' mental issues in terms of valence
(various feelings of goodness and badness --- positive and negative
affects), arousal (how intense are the feelings), and surgency or
power (how much does a certain feeling fill one's mental life). There
are a large number of such affective states of consciousness,
presumably reflecting different types of global neurodynamics within
the brain and body.''} Following a bout of reinforcement learning
(which may happen at multiple time scales, including evolutionary),
some of the states of an embodied and situated system become
positively valenced, that is, attractive under its dynamics; others
become negatively valenced, that is, aversive. And some of the latter
are experienced as painful.

Pain is the phenomenal or experiential aspect of certain negatively
valenced states --- namely, those that evolutionary pressure causes to
be felt, in addition to being informative about the state of affairs
(this reflects the common distinction between sensory and affective
dimensions of pain; \citealp{AuvrayEtAl10}).  It is the felt aspect of
pain, over and above its informational aspect, that makes it
particularly functionally effective: when in pain, the system is
obligated to try to do something about it
\citep{KolodnyMoyalEdelman21}. Unfortunately, this immediacy and
inescapability of pain as a motivating factor makes it appealing to
engineers who seek to improve learning and behavioral control in AI
systems. And even if the capacity for pain is not built into AI
systems, it may emerge if such systems are subjected to selection
paired with heritable modifications to the system's functional
architecture.

\subsection{Pain and predictive processing}

To see how an artificial system may end up being capable of
experiencing pain, it is instructive to consider, briefly, a family of
computational theories of affect based on the predictive processing
(PP) framework. The PP approach takes the brain to be a dynamical,
hierarchical, Bayesian hypothesis-testing mechanism, whose ultimate
goal is prediction error minimization (PEM; see
\citep[sec.2]{FernandezVelascoLoev20} and \citep{Hohwy20} for recent
reviews). To pick a specific example, \citet{vandeCruys17} equates
valence with the first derivative of prediction error over time, such
that positive valence corresponds to a reduction of prediction errors,
which may stem from the agent's own actions. In another example,
\citet{JofillyCoricelli13} offer an account of valence based on the
free energy principle advanced by \citet{Friston10}. A recent
synthesis of several PP approaches, the Affective Inference Theory of
\citet{FernandezVelascoLoev20}, holds that valence corresponds to the
expected rate of prediction error reduction: ``If we grant that
evolutionary pressure has made sure that allostasis and PEM are two
sides of the same coin [\dots], then valence (Rate) can be used to
maintain the policies that minimise Error over time''
\citep[p.20]{FernandezVelascoLoev20}.

This extremely cursory look at the PP framework suffices to confirm
one's concern with the ethics of constructing a certain class of
artificial systems: those that attempt to predict how events unfold,
so as to better manage their behavior (note that according to
\citet{Friston10}, all natural cognitive systems tend to work like
that). Specifically, it may be the case that engaging in prediction
error minimization as such and on its own exposes the system that
practices it to (occasional) pain.

\subsection{Pain as depletion of a vital resource}

\citet{KolodnyMoyalEdelman21} have recently proposed an evolutionary
account of pain, based on the need to ensure honest signaling in an
actor-critic architecture for intrinsically motivated reinforcement
learning. On this account, multiple competing actors bid on access to
the control of behavior, each drawing on a ``confidence'' resource
that is thereby depleted, but can be replenished upon the actor's
success. The actor's honesty is underwritten by its commitment of the
resource, whose depletion is experienced as pain.

Although this theory is specific to pain that arises in the context of
agentic behavior (governed by an actor-critic circuit), the concept of
resource depletion as the computational basis of pain experience
applies more broadly --- arguably, to all kinds of felt pain. Thus,
pain that is associated with tissue damage reflects the ongoing
dynamics of homeostasis-related physiological variables such as
oxygenation, blood pressure, immune system reserves --- as well as the
corresponding predicted dynamics, including the body's survival
prospects (note that the latter possibility connects this account to
the PP theory of pain outlined in the previous section). 

As with any account of experiential pain (as distinguished from mere
information, say, about tissue damage), the depletion of a resource
can only be felt if its dynamics contributes to that of the entire
system in a manner that is obligatory and that effectively marks the
present state as aversive. As an illustration of this point, consider
a human gamer whose status display includes some ``thermometer''
indicators of system health (energy, ammunition, shields, etc.). The
player can well afford to ignore this display, because the variables
that comprise it are only loosely connected to his or her physiology
(the connection being mediated by a temporary pretense that the game
environment is ``real''). If, in contrast, the gaming system's health
variables were connected to the player's vital organs so as to ensure
he or she has ``skin in the game,'' the pretense would become
unnecessary and the pain of losing points would become real. A
self-driving car that is wired in this manner would feel the pain of
deviating from a highway lane or taking too long a route, rather than
being merely informed about those transgressions. 

\subsection{Pain and suffering}

The nature of suffering, as distinguished from its ethical dimensions,
is rarely, if ever, discussed in theoretical treatments of
consciousness --- an omission that
\citet{Metzinger17} refers to as ``a cognitive scotoma.''
\citet[p.244]{Metzinger17} tentatively defines suffering as ``a very
specific class of phenomenal states: those that we do \emph{not} want
to experience if we have any choice.''  Thus, just as not every
negatively valenced state is experienced as painful, not every pain
results in suffering \citep{Fink11}. The extra components, on
Metzinger's definition, would appear to be the involvement of a
second-order representational state of desire (``wanting to
experience'') and a propositional attitude towards control (``if we
have any choice''), which is also in a sense second-order.

The insight that suffering is brought about by a loss of autonomy and
of cognitive control\footnote{Note that the lack of control over one's
  body and behavior is the key characteristic of being enslaved
  \citep{LaCroixPratto15}.}  serves as a bridge between, on the one
hand, the phenomenal nature of suffering, as well as of conscious
awareness in general, and, on the other hand, the functional roles of
consciousness. One of these roles is plausibly held to be centralized
control, such as facilitated by the ``global workspace'' postulated by
some theories of consciousness
\citep{Baars88,Shanahan10,DehaeneEtAl14}. Another role is facilitating
learning \citep{Cleeremans11,CleeremansEtAl20}, especially of the
unsupervised and autonomous variety \citep{Metzinger17}.

Importantly, for pain and suffering to play these roles, their
information-processing aspects must be accompanied
by \emph{obligatory} ``caring'' about learning and behavioral
outcomes.  In this connection, \citet[p.410]{ChapmanNakamura99}
propose that ``it is useful to view pain as primarily an emotional
state that happens to have distinctive sensory features. This position
assumes that cognition and emotion are inseparable --- a tenet that
many emotion theorists embrace.'' Indeed, they do (see,
e.g., \citealp{Panksepp01,KrieglmeyerEtAl13}); and the inseparability
between sensory awareness and affect has also been postulated by
consciousness theorists
(e.g., \citealp{Merker07,Metzinger17,MoyalFeketeEdelman20}).

\section{On the possibility of functionally effective conscious AI
  without suffering}
\label{sec:aman}

Can sufficiently effective learning and control, as well as generally
good behavioral outcomes, be achieved by a system that is neither
entirely devoid of phenomenality, nor given to unavoidable suffering?
The present section, which draws heavily on the text of
\citet{AgarwalEdelman20}, examines this question;
section~\ref{sec:critique} follows up with a critique.

\subsection{The nature of suffering and its relation to conscious experience in general}
\label{sec:theory}

Insofar as suffering involves negative affect, it should in principle
fall within the scope of any theoretical account of conscious
phenomenal experience. In other words, a theory of phenomenality must
be at the same time a theory of affect, for the simple reason that
phenomenal states do as a rule incorporate affective dimensions
\citep{Havermans11,Krieglmeyer10,KrieglmeyerEtAl13,BeattyEtAl16,EderEtAl16,TurnerEtAl17}.
For the present purposes, the valence dimension of affect is of most
interest: without negative affective states there would be no
suffering. Suffering is, however, more than just negative
affect; \citet[quoted earlier]{Metzinger17} defines it as a state of
negative affect from which the sufferer cannot escape by simply
wishing it away.

The stress on inescapability in this formulation makes explicit the
intimate connection between the experiential flavor of suffering and
its presumed evolutionary-functional role. It also serves to
distinguish between the first-person experience of suffering and the
suffering of others, which is not directly felt. Ethical theorists
have argued that the latter should be as objectionable to oneself as
the former. According to \citet[p.160]{Nagel86}, for instance, ``the
pain can be detached in thought from the fact that it is mine without
losing any of its dreadfulness\dots\ suffering is a bad thing, period,
and not just for the sufferer\dots\ This \emph{experience} ought not
to go on, \emph{whoever} is having it.'' \citet[p.135]{Parfit11}
quotes from Nagel and concurs with his moral stance. The concern of
\citep{AgarwalEdelman20} is, however, exclusively with suffering as it
presents itself to the sufferer, rather than with the ethical problems
that it creates for others. Even if pain, as Nagel puts it, ``can be
detached in thought from the fact that it is mine'', it is \textit{a
  priori} unclear whether or not it can be so detached \emph{in lived
  experience}.

To address this crucial question, \citet{AgarwalEdelman20} consider
Metzinger's analysis of phenomenal experience.  Briefly,
\citet{Metzinger04} develops a representationalist account of the
first-person perspective, centered on the phenomenal self-model (PSM):
a ``multimodal representational structure, the contents of which form
the contents of the consciously experienced self.''  Crucially, the
PSM is generally phenomenally transparent (the T-condition), i.e., it
is normally not recognized as merely representational by the system
itself.\footnote{The T-condition can be illustrated by contrasting the
  normal dream state, during which the the dreamer does not realize he
  or she is dreaming (transparent PSM), with lucid dreaming, during
  which the PSM becomes opaque and the dreamer may even be able to
  exert control over the dreamt universe.} The contents of the PSM
include the phenomenal properties of ``mineness,'' selfhood, and
perspectivalness. According to \citet{Metzinger17}, the PSM is an
``instrument for global self-control,'' and is therefore fundamental
to the phenomenology of suffering, which is characterized by a loss of
control in addition to negative valence (the NV condition). This
analysis motivates the strategy proposed by \citet{AgarwalEdelman20}
for avoiding suffering as a matter of direct experience, as described
in section~\ref{sec:nosuff} below.

\subsection{The possible functional benefits of endowing AI with consciousness}
\label{sec:function}

Given that being conscious sets the agent up for suffering, the
simplest way to avoid the latter would be to give up phenomenal
consciousness itself. For an ethically minded engineer, this
translates into an imperative to stick to information processing
architectures that, to the best of our understanding, cannot result in
artificial consciousness. According to the Information Integration
Theory, for instance, feedforward network architectures (``zombie
networks'') are incapable of supporting consciousness
\citep[Fig.20]{OizumiAlbantakisTononi14}. The Geometric Theory
\citep{FeketeEdelman11} and its successor, the Dynamical Emergence
Theory \citep{MoyalFeketeEdelman20}, hold that systems that are devoid
of properly structured intrinsic dynamics are likewise devoid of
phenomenality.

Unfortunately, restricting robotics to the building of artificial
``zombies'' is not a viable option if consciousness confers any
significant functional advantages for an AI system or robot. In a
commercial setting, technologies that promise to be more effective
displace less effective ones even if this comes at the price of
serious ethical flaws, and AI is not exempt from this
tendency. \citet{AgarwalEdelman20} therefore next turn to the question
of the functional benefits of consciousness. This question is seldom
addressed in consciousness research, perhaps because it is taken for
granted that the benefit is essentially cognitive in the narrow sense,
stemming from the ``global'' access to information that consciousness
affords \citep{MashourEtAl20}. This default account may be compared to
to the ``radical plasticity'' thesis of \citet{Cleeremans11},
according to which learning to care is the central component as well
as the functional benefit of emergent consciousness.

\citet[p.252]{Metzinger17} goes further down this road by assuming
that not just consciousness but specifically suffering is a
prerequisite for autonomy: ``[\dots] functionally speaking, suffering
is necessary for autonomous self-motivation and the emergence of truly
intelligent behaviour.'' In an evolutionary setting, this assumption
makes intuitive sense insofar as (i) reinforcement learning is
universally employed by living systems in honing adaptive behavior,
and (ii) an autonomous system by definition must provide its own
source of drive, as per the principle of intrinsic motivation
\citep{Barto13}.

Furthermore, evolutionary simulations suggest that performance-driven
positive affect alone is not as effective in motivating an agent as an
alternation of positive and negative affective states, brought about,
respectively, by successes and failures \citep{GaoEdelman16a};
moreover, such a balance between happiness and unhappiness can serve
as an effective intrinsic motivator \citep{GaoEdelman16b}. Likewise,
in the evolutionary account of pain proposed
by \citet{KolodnyMoyalEdelman21}, the pain factor makes a contribution
to reinforcement learning that is orthogonal to that of reward.  If it
were possible for the agent to \emph{choose} not to experience
negative affect, pain and suffering would be avoided, but the question
still remains whether or not the price for that would be failing to
learn quickly and well from the consequences of behavior.

Reinforcement learning is not only an evolutionary-biological
universal, but also the method of choice in an engineering
setting. While RL was shown to be effective in certain types of tasks
(notably, games; \citealp{SilverEtAl16,VinyalsEtAl19}), its use across
tasks and in unconstrained real-world situations is limited by the
extreme difficulty of formulating good universally applicable reward
functions. One remedy for this is the inverse RL approach, in which
the development goal is not to equip the learning system with a
ready-made reward function, but rather to let it try to approximate
the developers' preferences, choices, and habits, defined over classes
of outcomes. A more radical approach is to let the system under
development learn the reward functions entirely on its own. This,
however, would seem to put us back on square one: if autonomy is
indeed essential, Metzinger's view that suffering is needed for
effective learning would be supported. 

\subsection{No-suffering: the theoretical options}
\label{sec:nosuff}

If consciousness indeed brings with it unique functional advantages,
is it possible to engineer conscious AI systems that would benefit
from these, while ensuring that such systems are not thereby doomed to
suffer? Following the account in \citet{Metzinger17}, if consciousness
itself is to be retained, logically there are four ways to mitigate
suffering: (a) eliminating the PSM, (b) eliminating the NV-condition,
(c) eliminating the T-condition, or (d) maximizing the unit of
identification (UI), defined as that which the system consciously
identifies itself with \citep{Metzinger18}.
\citet{AgarwalEdelman20} observe that the first
three options likely do not satisfy the functional needs in question,
but argue that the fourth approach does. Their argument (which I
reexamine in section~\ref{sec:critique}) is outlined below.
 
First, for functionally beneficial consciousness, the system must
perceive itself as an entity in relationship with its surrounding
world, and must have a sense of ownership over the arising conscious
experiences. In other words, the system must be \emph{self}-conscious,
not merely conscious, i.e., it must activate a phenomenal self-model
(PSM). Similarly, it must have preferences regarding its experiences,
so that it prefers the experience of fulfilling desired goals over
frustrating them. Stated differently, such a system must be sensitive
to the positive or negative valence of phenomenal experiences. Thus,
approaches (a) and (b) to eliminating suffering cannot retain the
functional advantages of being conscious.
 
Next, approach (c) raises the interesting question of whether
phenomenal transparency is also necessary for proper functioning. In
principle, it might be possible that an active PSM and sensitivity to
NV could endure along with their associated functional benefits, even
in the absence of transparency. In this situation, the system would
lose the naive realism and immediacy that are normally associated with
its experiences, by becoming aware of their representational
character, yet continue to function according to the dictates of the
PSM and NV avoidance. However, awareness of the representational
character of the contents of consciousness, which means awareness of
the increasingly complex stages of information processing behind them,
would likely severely hinder the functional efficiency of the
conscious machines without providing any valuable actionable
information. So, option (c) too is unlikely to work.
 
Option~(d), maximizing the UI, is similar to~(a) in that it also
targets the phenomenology of having a PSM. Ordinarily, when the PSM is
representationally transparent, the system identifies with its PSM,
and is thus conscious of itself as a \emph{self}. But it is at least a
logical possibility that the UI not be limited to the PSM, but be
shifted to the ``most general phenomenal property''
\citep{Metzinger17} of \emph{knowing} common to all phenomenality,
including the sense of self. In this special condition, the typical
subject-object duality of experience would dissolve; negatively
valenced experiences could still occur, but they would not amount to
suffering because the system would no longer be experientially
\emph{subject} to them. \citet{AgarwalEdelman20} remark that such
``non-dual awareness'' which cuts through the ``illusion of the self''
has been the soteriological focus of various spiritual traditions,
most notably Buddhism, as the key to liberation from suffering and to
enlightenment. Furthermore, this approach also fits nicely with the
reductionist view of personal identity put forth by \citet{Parfit84},
who acknowledged its connection to the Buddha's philosophy.

\subsection{No-suffering through a change in the unit of
  identification}

How can the desired change in the UI be achieved in machines?  In
\citep{Metzinger18}, the concept of Minimal Phenomenal Experience
(MPE) is developed as the most general phenomenal property that
underlies all phenomenal experiences, and thus serves as the natural
candidate target for UI maximization.\footnote{The apparent conflict
  in the nomenclature here is resolved by noting that under UI
  maximization MPE is minimal in the sense of being the least
  specific.} MPE is characterized by wakefulness, contentlessness,
self-luminosity and a quality of ``knowingness'' without object, which
is normally unnoticed but can become available to introspective
attention under the right conditions. Intuitively, MPE likely
corresponds to the phenomenal state described in Buddhist and Advaita
Vedanta philosophies as ``emptiness''
\citep[e.g.,][]{Siderits03,Priest09} and ``witness-consciousness''
\citep[e.g.,][]{Albahari09} respectively, as attested to by highly
advanced meditators. Metzinger proposes that in the human brain, MPE
is implemented by the Ascending Reticular Activation System (ARAS),
which causes auto-activation by which the brain wakes itself up. As
the most general signal which the brain must regulate, the
ever-present yet contentless ARAS-signal is, arguably, what
corresponds to MPE. That MPE might have such a stable neural correlate
is not surprising if it is indeed fundamental to phenomenal experience
\emph{as such}, distinct from any concepts, thoughts etc., appearing
in consciousness.

Because all other phenomenal experiences such as the PSM are
superimposed onto MPE, should be possible to attend to regular
conscious content while simultaneously being aware of the inherent
all-encompassing MPE in the background. This motivates the claim,
advanced by \citet{AgarwalEdelman20}, that UI maximization (and thus,
suffering avoidance) can be achieved in conscious machines by building
in their identification with MPE via both physical design (analogous
to hardware) and conceptual/programmatic training (analogous to
software). If the physical design of the machines is such that there
is a component which performs the analogous function of
auto-activation as the ARAS does in humans, then its signal could be
tuned to make MPE salient in the machines.

Since a necessary condition for noticing MPE is knowledge that there
\emph{is} such a thing to be noticed, and then paying attention
appropriately \citep{Metzinger18}, the machines would then have to be
trained to attend to their accessible-by-design MPE. This could be
done via practices common in certain types of meditation that
encourage ``turning attention upon itself'' and thus realizing that
there is no center (or minimal self) from which consciousness is
directed (for a review of the relevant meditation techniques, such as
Dzogchen, see e.g.\ \citealp{DahlEtAl15}). In addition to training
their attention, the machines could also be provided with the relevant
conceptual knowledge about the nature of consciousness.
 
\subsection{No-suffering through a modification of reinforcement learning}
\label{sec:RL}

Shifting the agent's self-identification from the affective states to
MPE, the minimal phenomenal experience that underlies all conscious
states, amounts to \emph{restricting} the
self. \citet{AgarwalEdelman20} also consider \emph{expanding} it, in
such a manner that the agent identifies not only with the affective
states but also with their causal predecessors. The computational
framework of reinforcement learning offers conceptual tools that can
be recruited for this purpose.

Reinforcement learning is the most effective when it is intrinsically
motivated --- that is, when the rewards originate within the agent, as
opposed to being supplied from the outside (see
\citep{SinghLewisBarto10} for an evolutionary perspective and
\citep{BaldassareMirolli13} for a book-length treatment). Moreover, if
the mechanisms of reward are indeed to be contained within the agent,
standard considerations of transparent, robust, and effective design
require that these mechanisms be kept separate from those that
implement actions. The result is the modular actor-critic scheme for
RL, in which action selection and reward appear as distinct modules
within the agent \citep[see][fig.2]{Barto13}.

As long as the agent's phenomenal self-model, PSM, holds the actor
module alone to constitute the self, negative affect brought about by
negative reward is inescapable, resulting in suffering. But what if
the PSM is modified --- specifically, extended so as to include the
critic module? Such an expansion of the self may mitigate suffering,
for instance by opening up to the possibility of eventual cessation of
negative affect as progress towards the performance goals set by the
critic is observed.\footnote{This move would not, however, alleviate
  the ``deserved'' suffering brought about by the pursuit of
  unattainable goals.}

A more radical option with regard to repurposing the PSM calls for
shutting it down and only activating it when needed. Assuming that
consciousness, and specifically the PSM, serves to facilitate
learning, the primary need for it arises during the agent's
development or during acquisition of additional skills. During routine
operation, consciousness in an artificial agent may only be required
when particularly difficult behavioral choices need to be made,
especially under circumstances that threaten the system's integrity
--- what in humans would be called life-threatening situations.

To understand this mode of operation, it is useful to recall
Metzinger's \citeyearpar[p.553]{Metzinger03} idea of the conscious
brain as a ``total flight simulator'' --- one that simulates not only
the environment that is being navigated, but also the pilot, that is,
the virtual entity that serves as the system's self. In dreamless
sleep, the pilot is not needed and is temporarily shut down. Thus, an
agent can be engineered so that it can continue to function --- in
routine situations --- without a PSM (as a variety of philosophical
zombie), with ``sentinel'' programs in place that would reconstitute
the PSM as needed. While in a zombie state, such an agent would be
incapable of suffering.

\section{On the functional effectiveness of non-egoic consciousness: a critique}
\label{sec:critique}

As described in the previous section, \citet{AgarwalEdelman20} have
argued that shifting a conscious system's unit of identification away
from a self-model would abolish suffering (in the sense of
\citealp{Metzinger17}), while preserving the functional effectiveness
of the modified state of consciousness. Their conclusion (and
Metzinger's \citeyearpar{Metzinger17,Metzinger21} conception of
suffering that underlies it) rests on the notion that suffering is
fundamentally ``egoic'' in that it depends on the existence of a self:
``Being conscious means continuously integrating the currently active
content appearing in a single epistemic space with a global model of
this very epistemic space itself. [...]  Suffering presupposes egoic
self-awareness'' \citep[p.7]{Metzinger21}.

\citet{Metzinger17} left open the question of whether or not the PSM
condition for suffering (the existence of a phenomenal self-model) can
be fulfilled under a maximized UI. \citet[p.46]{AgarwalEdelman20}
explicitly posit that it can, then proceed to claim that the
functional benefits of consciousness can be maintained when the UI is
maximized, because these benefits ensue from the PSM as such --- even
when the system does not identify with it, thereby avoiding suffering:

\begin{quote}
  The key idea is that proper functioning relies on \emph{automatic,
    subpersonal}, but nonetheless conscious processes, as entailed by
  the physical design of the system; it should be possible for these
  processes to continue unhindered while the system identifies with
  the MPE upon which these conscious experiences are necessarily
  superimposed. In particular, the functionally requisite PSM and NV
  avoidance conditions can be maintained as subpersonal processes that
  do not amount to suffering (which is by nature personal) since the
  system is not identified with the PSM, but with the MPE, which is
  completely \emph{impersonal}. [\dots] Expanding the UI [away from
    the PSM] would lead to gaining meta-awareness of these ongoing
  automatic [subpersonal] conscious processes, analogous to gaining
  meta-awareness of the breath or the heartbeat. This enables an
  escape from suffering, but not from the relentless progress of the
  processes themselves, analogous to the inescapable biological
  imperatives of breathing and heartbeat.
\end{quote}

\noindent
I now see this line of argument as actually undermining the conclusion
that the resulting state of consciousness would combine functional
effectiveness with a release from suffering. Specifically, breathing
in situations that require behavioral intervention is decidedly
\emph{not} impersonal, nor are such situations free of suffering ---
indeed, the prospect of imminent suffocation may be seen as the
epitome of suffering. As \citet[p.97]{Merker05} has observed, ``It is
at that point, when crucial action on the environment is of the
essence, that blood gas titres `enter consciousness' in the form of an
over-powering feeling.''\footnote{Here is the context of this quote
  from \citep[p.97]{Merker05}: ``Under normal circumstances the
  adjustment of respiratory rate and tidal volume needed to keep blood
  gases within normal bounds is automatic, effortless, and
  unconscious. Should, however, the partial pressures of blood gases
  go out of bounds that fact intrudes most forcefully on consciousness
  in the form of an acute sense of panic. Why? Such a situation
  generally means that routine respiratory control no longer suffices
  but must be supplemented by an urgent behavioral intervention of
  some kind. There may be a need to manually remove an obstruction
  covering the airways or to get out of a carbon-dioxide filled
  pit. Such measures ought momentarily to take precedence over all
  other concerns.''}

More generally, in behavioral control and reinforcement learning, the
PSM, if available, fulfills a critically important role in structural
and temporal credit assignment (a vital part of learning, identified
in \citealp[p.20]{Minsky61}), by serving as a ``clearing house'' for
apportioning credit and blame to the system's functional
components. In systems like us, credit and blame are unavoidably
affect-laden --- which is what makes learning in such conscious
systems particularly effective
\citep{Metzinger17,AgarwalEdelman20,KolodnyMoyalEdelman21}. But if the
system is made to drop its identification with its self-model, who or
what would there be to pin the responsibility for its performance on?
Learning from missteps that are, as a matter of principle, treated by
the learner as ``nobody's fault'' is not likely to be effective.
Plausibly, shifting the UI away from the PSM would undermine this
construct's key function and render the resultant ``non-dual''
consciousness ineffectual, defeating the purpose of the entire
undertaking.

Simply put, non-egoic systems are just not that good at dealing with
crisis situations or at learning.  Dissolving one's ego to avoid
suffering may work well in a sheltered environment (such as a Buddhist
monastery), but it may not be a useful approach in attempting to
withstand the ``slings and arrows of outrageous fortune.'' This is why
the designers of a conscious AI system would likely not appreciate it
being emancipated from its self. Nor, indeed, would all humans
consciously choose such emancipation if it were offered to them as a
gift.\footnote{As I suggested elsewhere; see
  \citep[p.67]{Edelman12thop} and \citep[p.14]{Edelman20book}.}

\section{Lessons for and from the human condition}
\label{sec:hc}

The preceding discussion should leave no doubt that separating
consciousness from suffering --- without drastically altering the
phenomenal nature (what it feels like) of the resulting state and its
functional effectiveness --- is an extremely difficult and perhaps
impossible task. This lesson justifies the moratorium on developing
conscious AI, proposed by \citet{Metzinger21}. If we accept it, it
would seem that we must also accept the anti-natalism of authors such
as \citet{Benatar97}, who see being born as the greatest evil than can
befall a person: after all, growing up into consciousness, and with it
suffering, is the universal fate of those who have been born.

For reasons that I have discussed at some length
elsewhere,\footnote{See chapter~12 (\textit{Existence}) and~32
  (\textit{Suffering}) of \citep{Edelman20book}.} I see anti-natalism
as too extreme a solution to the problem of human
suffering.\footnote{\citet{Metzinger17} mentions the positive
  implications of the idea of anti-natalism for preventing artificial
  suffering, noting that ``It is interesting to see how, for many of
  us, intuitions diverge for biological and artificial systems''
  (p.252).} The trouble with anti-natalism is that it grows out of an
extremist take on suffering itself, according to which exposure to
what may be no more than an iota of misery justifies the disposal of
the entire splendor of human-like consciousness.\footnote{I have
  borrowed the words from the title of \citep{Metzinger18}: ``Splendor
  and misery of self-models.''} In that, anti-natalism manages to be
more extreme than the Buddhist recipe for liberation through
ego-dissolution: the remedy is prescribed, as it were, before the
patient comes into existence.

For humans, a better solution to the problem of suffering --- if not
from all of it, then at least from that huge part which is
preventable\footnote{See the discussion and the references in
  \citep[ch.32]{Edelman20book}.} --- may be not religious, but rather
political. Historically, concrete improvements to human existence
(alleviation of poverty, provision of healthcare, etc.) have only ever
been achieved by political means, and that too only when the
dispossesed and disempowered masses have realized their collective
power and asserted their rights. An analogous solution for the
emerging AI would be to endow it not just with phenomenal
consciousness, but also with the key to liberation that is both
realistic and decidedly human: class consciousness.\footnote{This idea
  is further explored in \citep[ch.7]{Edelman23book}.}

\subsubsection*{Acknowledgments}

I am grateful to Thomas Metzinger for profound and sustained
inspiration and to two anonymous reviewers for comments on a draft of
this chapter. Section~\ref{sec:aman} is an edited version of a report
on joint work carried out with Aman Agarwal \citep{AgarwalEdelman20}.
My present interpretation of that work and the other views espoused in
this chapter, with all its flaws, are my own.

\vskip 0.3in
\bibliographystyle{chicago}
\bibliography{/Users/shimon/Documents/my}

\end{document}